\documentclass[
twocolumn,
]{ceurart}

\usepackage{listings}
\usepackage{times}
\usepackage{soul}
\usepackage{url}
\usepackage{graphicx}
\usepackage{amsmath}
\usepackage{amsthm}
\usepackage{booktabs}
\usepackage{algorithm}
\usepackage{algorithmic}
\usepackage[switch]{lineno}
\hypersetup{colorlinks=true}
\usepackage{subcaption}

\usepackage{array}
\usepackage{graphicx}
\usepackage{booktabs} 
\usepackage{multirow} 
\usepackage{tabularx}
\usepackage{natbib}
\usepackage{amsfonts}
\usepackage{comment}
\begin{document}

\copyrightyear{2024}
\copyrightclause{Copyright for this paper by its authors.
 Use permitted under Creative Commons License Attribution 4.0
 International (CC BY 4.0).}

\conference{The IJCAI-2024 AISafety Workshop}

\title{Detecting out-of-distribution text using topological features of transformer-based language models}


\author[1]{Andres Pollano}[%
email=apollano@student.unimelb.edu.au,
]
\address[1]{University of Melbourne, Melbourne, Australia}

\author[2]{Anupam Chaudhuri}[%
email=anupam.chaudhuri@deakin.edu.au,
]
\cormark[1]
\address[2]{Deakin University, Geelong, Australia}

\author[3]{Anj Simmons}[%
email=anj@simmons.ai ,
]
\address[3]{Hashtag AI,
Melbourne, Australia}

\cortext[1]{Corresponding author.}

\begin{abstract}
  To safeguard machine learning systems that operate on textual data against out-of-distribution (OOD) inputs that could cause unpredictable behaviour, we explore the use of topological features of self-attention maps from transformer-based language models to detect when input text is out of distribution. Self-attention forms the core of transformer-based language models, dynamically assigning vectors to words based on context, thus in theory our methodology is applicable to any transformer-based language model with multihead self-attention. We evaluate our approach on BERT and compare it to a traditional OOD approach using CLS embeddings. Our results show that our approach outperforms CLS embeddings in distinguishing in-distribution samples from far-out-of-domain samples, but struggles with near or same-domain datasets.
\end{abstract}


\maketitle

\section{Introduction}

Machine learning (ML) models perform well on the datasets they have been trained on, but can behave unreliably when tested on data that is out-of-distribution (OOD). For example, when a ML model has been trained to recognise different breeds of cats is fed an image of a dog, the results are unpredictable. OOD detection is the task of identifying that an input does not seem to be drawn from the same distribution as the training data, and thus the prediction given by the ML model should not be trusted. OOD detectors can be used to defend ML models deployed in high stakes applications from OOD data by providing a warning/error message for OOD inputs rather than processing the input and producing untrustworthy results \cite{wong2023mlguard}.

In this paper, we focus on OOD detection for textual inputs to safeguard ML models that perform natural language processing (NLP) tasks. For example, a sentiment classification model trained on formal restaurant reviews may not produce valid results when applied to informal posts from social media. Determining that an input is OOD requires a way to measure the distance between an input and the in-distribution data. This in turn requires a method to convert textual data into an embedding space in which we can measure distance. One approach to this is to input the text to a transformer-based language model, such as BERT \cite{devlin-etal-2019-bert}, to extract an embedding vector for the input text (e.g., the hidden representation of the special $[CLS]$ token). We can then measure the distance of the embedding vector for an input text to the nearest (or k-nearest) embedding vector of a text from an in-distribution validation set. When this distance is beyond some threshold (which needs to be calibrated for the application), the input text is flagged as out of distribution.
The internal state of transformer-based language models contains important information, which may be able to offer richer representations than only using the embedding obtained from the last or penultimate layer. For example, \citet{azaria2023internal} demonstrated that it is possible to train a classifier on the activation values of the hidden layers of large language models to predict when they are generating false information rather than true information. However, training a classifier for OOD detection in this manner is not a suitable approach, as the distribution of the OOD data that will be encountered is not knowable in advance. That is, due to the nature of OOD detection, we need to extract an embedding vector and associated distance metric (calibrated solely on the training/validation data) without training a further classifier over this space.

Recently, \citet{DBLP:journals/corr/abs-2109-04825} proposed an approach to analyze the topology of attention maps of transformer-based language models to determine when text had been artificially generated, and \citet{perez2022topological} propose using the topology of attention maps of transformer-based language models to detect adversarial textual attacks. Specifically, topological data analysis (TDA) provides a way to extract high-level features (related to the topology of the attention maps for each attention head in each layer) that can serve as an embedding vector of lower dimension than the full internal model state. In this paper, we investigate the suitability of these topological embeddings for the task of OOD detection, and contrast them to traditional approaches. Some of the work related to out-of-distribution detection in the context of transformer-based language models and using Mahalanobis distance can be referred to here~\cite{podolskiy2021revisiting, colombo2022beyond, li2021k, lee2018simple}.

We have made the code 
used to generate our results public under the MIT licence, with the intention of aiding the application of TDA methods to transformer-based models.
\footnote{\url{https://github.com/andrespollano/neural_nets-tda}}

\section{Background}
\subsection{ Topological Data Analysis}

Topology studies properties of geometric objects invariant under continuous deformation. For instance, a donut and a coffee cup are topologically equivalent. Algebraic topology, as in Hatcher's work~\cite{allen2002cambridge}, attaches algebraic objects such as groups to topological spaces. Certain features of these algebraic object can help to quantify those topological spaces. 

Persistence extends topology to finite data sets, tracing back to~\citet{frosini1992measuring, robins1999towards}. Persistence homology groups, derived from homology groups, serve as invariants for discrete objects.

For any finite set of points, we can construct a distance matrix where both the rows and columns are labeled by these points, and each entry in the matrix represents the distance between a pair of points. We can apply tools from Topological Data Analysis (TDA) to this set of points, allowing us to assign certain invariant characteristics to the collection.

In the context of language or text, we can think of each word as a point in some vector space, with a distance defined between words. For example, the distance might be related to semantic similarity or other linguistic relationships. By considering a text as a collection of such points, we can assign various numerical characteristics to it. These characteristics can distinguish the text from others and provide insights into its structure and content.

\subsubsection{Simplicial Complex and Chain}

A \textbf{simplicial complex} is a fundamental construct in algebraic topology, used to approximate and study more complex topological spaces. It is formed by combining simpler building blocks called simplices.

\textbf{Simplices:} A $k$-dimensional simplex, denoted as $\sigma$, is the convex hull of $k+1$ affinely independent points. For example, a 0-simplex is a point, a 1-simplex is a line segment, a 2-simplex is a triangle, and a 3-simplex is a tetrahedron.

\textbf{Forming a Simplicial Complex:} A simplicial complex $K$ in $\mathbb{R}^d$ is a collection of simplices that satisfies two conditions:
\begin{enumerate}
    \item Any face of a simplex in $K$ is also in $K$.
    \item The intersection of any two simplices in $K$ is either empty or a common face of both.
\end{enumerate}

\textbf{Simplicial Chains:} To study the algebraic properties of simplicial complexes, we introduce the concept of simplicial chains. A simplicial chain in a complex is a formal sum of simplices. For a given dimension $k$, the group of $k$-chains, denoted $C_k$, is the free abelian group generated by the $k$-dimensional simplices of the complex.

\textbf{Boundary Operators:} The boundary of a simplex is the sum of its faces. The boundary operator $\partial_k : C_k \to C_{k-1}$ maps each $k$-simplex to its $(k-1)$-dimensional boundary. This operator is crucial for defining the homology of the complex.

For example, the boundary of a 2-simplex (triangle) $\sigma = [v_0, v_1, v_2]$ is the sum of its 1-dimensional faces (edges): $\partial_2(\sigma) = [v_1, v_2] + [v_2, v_0] + [v_0, v_1]$.

\textbf{Chain Complex:} A chain complex is a sequence of chain groups connected by boundary operators:
\begin{align*}
0 \rightarrow C_n \xrightarrow{\partial_{n}} C_{n-1} \xrightarrow{\partial_{n-1}} \cdots \rightarrow C_1 \xrightarrow{\partial_1} C_0 \rightarrow 0.
\end{align*}

\textbf{Cycle and Boundary Groups:}
\begin{align*}
Z_p &= \text{ker } \partial_p, \quad B_p = \text{im } \partial_{p+1}, \quad B_p \subset Z_p.
\end{align*}

\textbf{Simplicial Homology:} The $k^{\text{th}}$ simplicial homology group of a complex $K$ is $H_k(K) = Z_k(K) / B_k(K)$, with the Betti number $\beta_k(K) = \dim H_k(K)$.

\subsubsection{Vietoris-Rips Complex}

The Vietoris-Rips complex is a key construct in topological data analysis, used for forming a simplicial complex from a set of data points based on their pairwise distances.

\textbf{Definition:} Given a set of points $X$ and a distance threshold $\varepsilon$, the Vietoris-Rips complex $\mathcal{VR}_\varepsilon(X)$ is defined as follows: for any subset $\sigma \subseteq X$, $\sigma$ is a simplex in $\mathcal{VR}_\varepsilon(X)$ if and only if the distance between every pair of points in $\sigma$ is less than or equal to $\varepsilon$.

\textbf{Formal Construction:}
\begin{itemize}
    \item \textit{Vertices:} Each point in $X$ is a 0-simplex (vertex).
    \item \textit{Edges:} An edge (1-simplex) connects vertices $x_i$ and $x_j$ if $d(x_i, x_j) \leq \varepsilon$.
    \item \textit{Higher Simplices:} A $k$-simplex is formed by a set of $k+1$ vertices if every pair of vertices in the set is connected by an edge.
\end{itemize}

\subsection{BERT Model}
BERT \cite{devlin-etal-2019-bert} is a transformer-based language model that has been pre-trained on a large corpus of text from BooksCorpus and English Wikipedia. Input text first needs to be tokenized, in which each word is converted to one or more tokens. The first token is the special $[CLS]$ token, followed by the tokenization of each word, using the special $[SEP]$ token to separate ``sentences'' (e.g., question and answer, these don't necessarily correspond to linguistic sentences). BERT is trained to achieve two objectives: Masked Language Modelling (MLM) in which tokens are masked at random (replaced with the special $[MASK]$ token) and the language model needs to learn to fill these in; and Next Sentence Prediction (NSP) in which the final hidden vector of the special $[CLS]$ token is used to predict if two sentences follow each other in the corpus.

As a transformer-based model, BERT consists of multiple layers, each with multiple attention heads. While multiple variants of BERT are available, for the purpose of this paper we use $BERT_{BASE}$, which consists of 12 layers, each with 12 attention heads (i.e., 144 attention heads in total) that operate on an input matrix, $X$, of $n$ tokens and 768 hidden dimensions, $d$.
\subsubsection{Sentence Embeddings}

The final hidden vector of the special $[CLS]$ token can be used to embed the input sequence (which varies in length) in $d$ hidden dimensions (178 in the case of $BERT_{BASE}$). The authors of the BERT paper \cite{devlin-etal-2019-bert} note that the $[CLS]$ embedding is not a meaningful sentence representation without fine-tuning. Nevertheless, \citet{uppaal-etal-2023-fine} claim that the practice of using this to obtain sentence embeddings ``is standard for most BERT-like models'', and find that in the case of RoBERTa (a BERT-like model without the NSP training objective) this embedding serves as a ``near perfect'' OOD detector even without fine-tuning.

\subsubsection{Attention Maps}

Each attention head computes an attention map, $W^{attn}$, of shape $n \times n$ as an intermediate step of the calculation. We use the same definition of attention maps as \citet{DBLP:journals/corr/abs-2109-04825} presented below:

\begin{align*}
X^{out} &= W^{attn}(XW^{V}) \\
W^{attn} &= \text{softmax} \left(\frac{(XW^Q)(XW^K)^T}{\sqrt{d}}\right)
\end{align*}

Where $W^Q$, $W^K$, $W^V$ are learned projection matrices of shape $d \times d$ and $X^{out}$ is the output of the attention head applied to the $n \times d$ matrix $X$ from the previous layer. In this paper, we analyse the attention maps for each of the 144 attention heads in $BERT_{BASE}$ using TDA.

\section{Experiment design}
In this section, we outline the design of our methodology for our OOD detection using Topological Data Analysis. For a supervised classification task, given a test sample $x$, OOD detection aims to determine whether it belongs to the in-distribution (ID) dataset $x \in \mathcal{D}_{in}$ or not. 
Some of the background and literature review related to confidence score for OOD detection can be found in~\cite{lee2018simple, sun2022out, yang2021generalized}.
We consider a $d$-dimensional representation of an input text $x$ as $h(x)$ in $\mathbb{R}^d$. To analyse the benefits of TDA in OOD detection, we consider two encoding functions $h_1(x)$ and $h_2(x)$:

\begin{enumerate}
    \item Topological feature vector $h_1(x)$: given $x$, we generate a vector of $d_1$ topological features using the graph representations of the 144 attention maps generated by $BERT_{BASE}$. In \ref{sec:attention-maps} and \autoref{sec:persistent-homology}, we explain in detail how the topological features are generated from an input sentence.
    \item Sentence embedding $h_2(x)$: we take the $d_2$-dimensional text embedding of the $[CLS]$ token output by $BERT_{BASE}$, which captures the contextual and semantic information of the input text $x$.
\end{enumerate}

Similar to \citet{uppaal-etal-2023-fine}, we define the OOD detection function as $G(x)$, which maps an instance $x$ to $\{in, out\}$ as follows:

\[G_{\lambda}(x; h) = \begin{cases}
in & \text{if } S(x;h) \ge \lambda\\
out & \text{if } S(x;h) < \lambda
\end{cases}\]

where $S(x;h)$ is an OOD scoring function using a distance-based method (Mahalanobis distance to the ID class centroids or Euclidean distance to k-nearest ID neighbour), described in \autoref{sec:ood-scoring}, and $\lambda$ is the threshold chosen so that a high proportion of ID samples’ scores are above $\lambda$. 

\subsection{Data}
As the in-distribution dataset, we choose the headlines and abstract text of ‘Politics’ and ‘Entertainment’  news articles from HuffPost from the \textit{news-category} dataset \cite{misra2022news}. To test the robustness of the OOD method, we conduct experiments on three kinds of dataset distribution shifts \cite{Arora2021TypesOO}:

\begin{itemize}
    \item \textbf{Near Out-of-Domain shift}. In this paradigm, ID and OOD samples come from different distributions (datasets) exhibiting semantic similarities. In our experiments, we evaluate the abstract of news articles from the \textit{cnn-dailymail} dataset \cite{see-etal-2017-get}.
    \item \textbf{Far Out-of-Domain shift}. In this type of shift, the OOD samples come from a different domain and exhibit significant semantic differences. In particular, we evaluate the IMDB movie review dataset \cite{maas-EtAl:2011:ACL-HLT2011} as OOD samples. 
    \item \textbf{Same-Domain shift}. We also test a more challenging setting, where ID and OOD samples are drawn from the same domain, but with different labels. Specifically, we extract the ‘Business’ news articles from the \textit{news-category} dataset.
\end{itemize}

In our experiments we used a sample of 30,000 points from the in-distribution dataset for the fine-tuned version of the model, and use a validation and test size of 1,000 datapoints.

\subsection{Model}
We focus on the attention heads of a pre-trained $BERT_{BASE} $ (L=12, H=12) generated from an input text $x$ to produce topological features and compare this encoding to the embeddings of the $[CLS]$ token as the sentence representation. We replicate our experiments on a fine-tuned $BERT_{BASE}$ on the ID news categorisation task $\mathcal{X} \rightarrow {\{\text{'Politics'}, \text{'Entertainment'} \}}$. We fine-tune the model for 3 epochs, using Adam with batch size of 32 and learning rate $10^{-5}$.

\subsection{Attention Maps and Attention Graphs}
\label{sec:attention-maps}

\begin{figure}[ht!]
    \centering
    \begin{subfigure}[t]{0.2\textwidth}
        \centering
        \includegraphics[width=\linewidth]{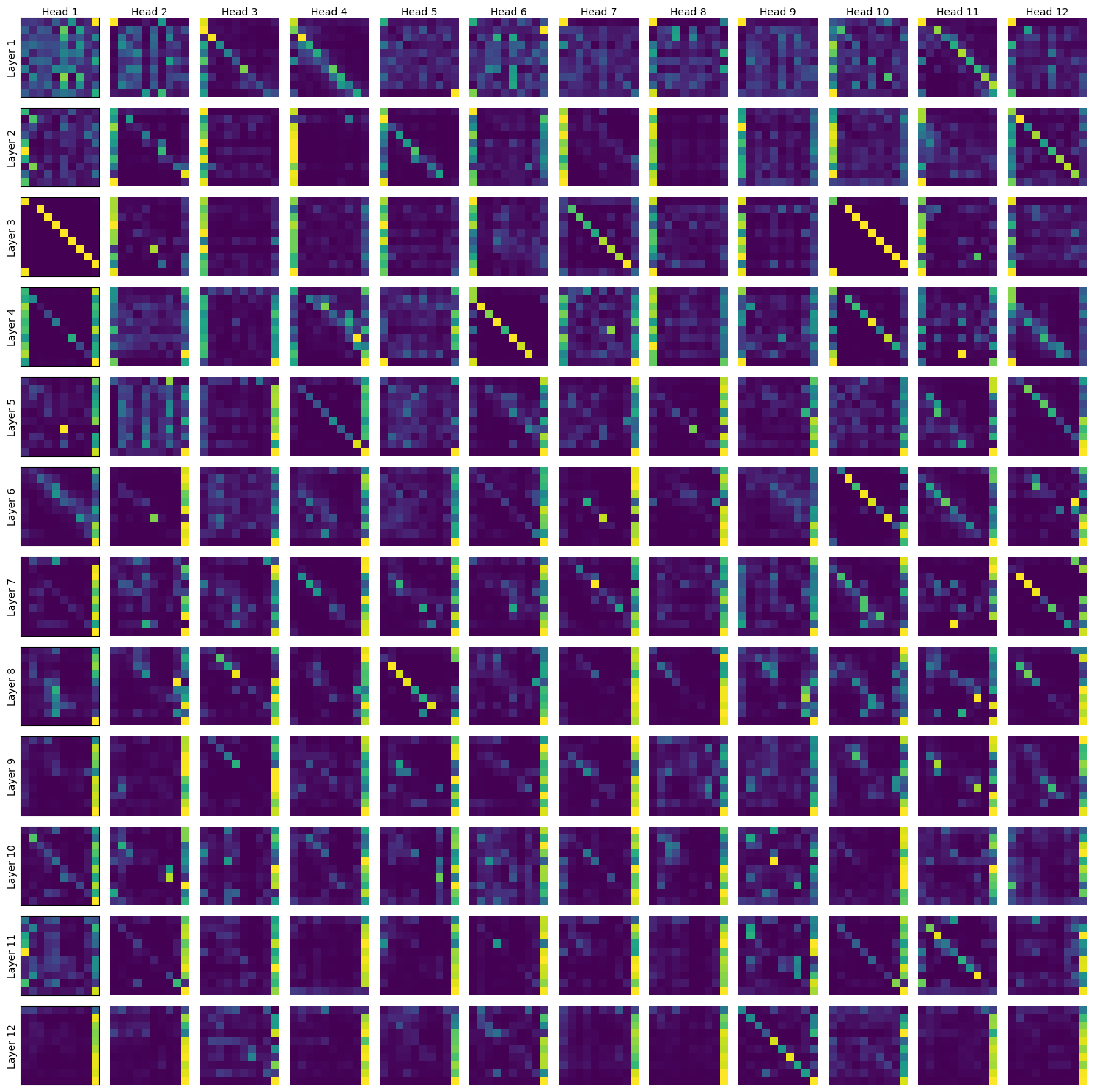}
        \caption{Attention maps $\left( 12 \times 12 \right)$ derived from pre-trained BERT for the input text "President issues vows as tensions with China rise"}
        \label{fig1a}
    \end{subfigure}%
    \quad
    \begin{subfigure}[t]{0.2\textwidth}
        \centering
        \includegraphics[width=\linewidth ]{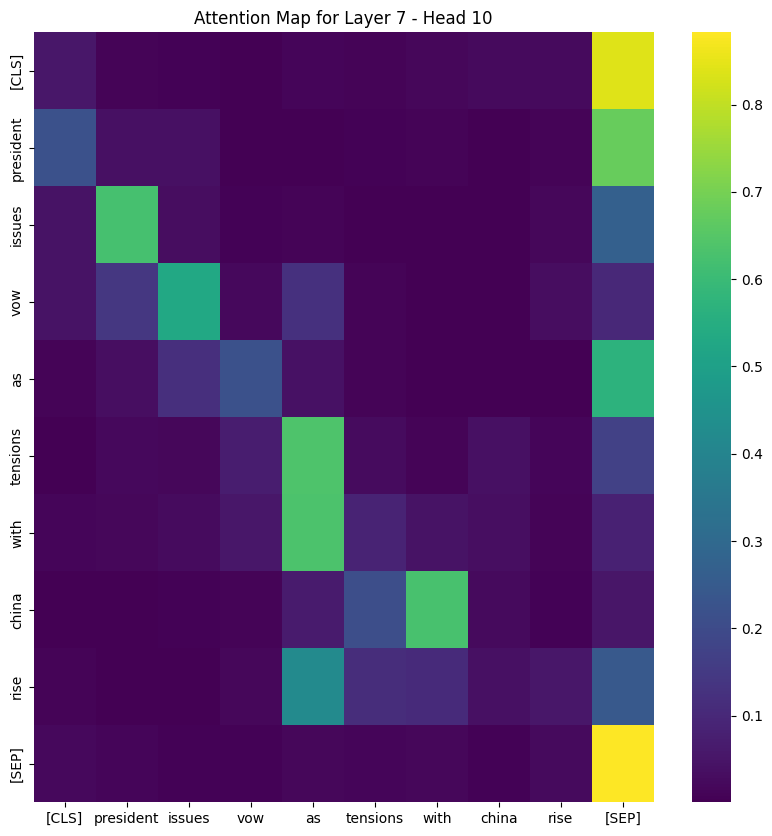}
        \caption{BERT Attention Map \\ (Layer 7; Head 10)}
        \label{fig1b}
    \end{subfigure}%
    \begin{subfigure}[t]{0.2\textwidth}
        \centering
        \includegraphics[width=\linewidth]{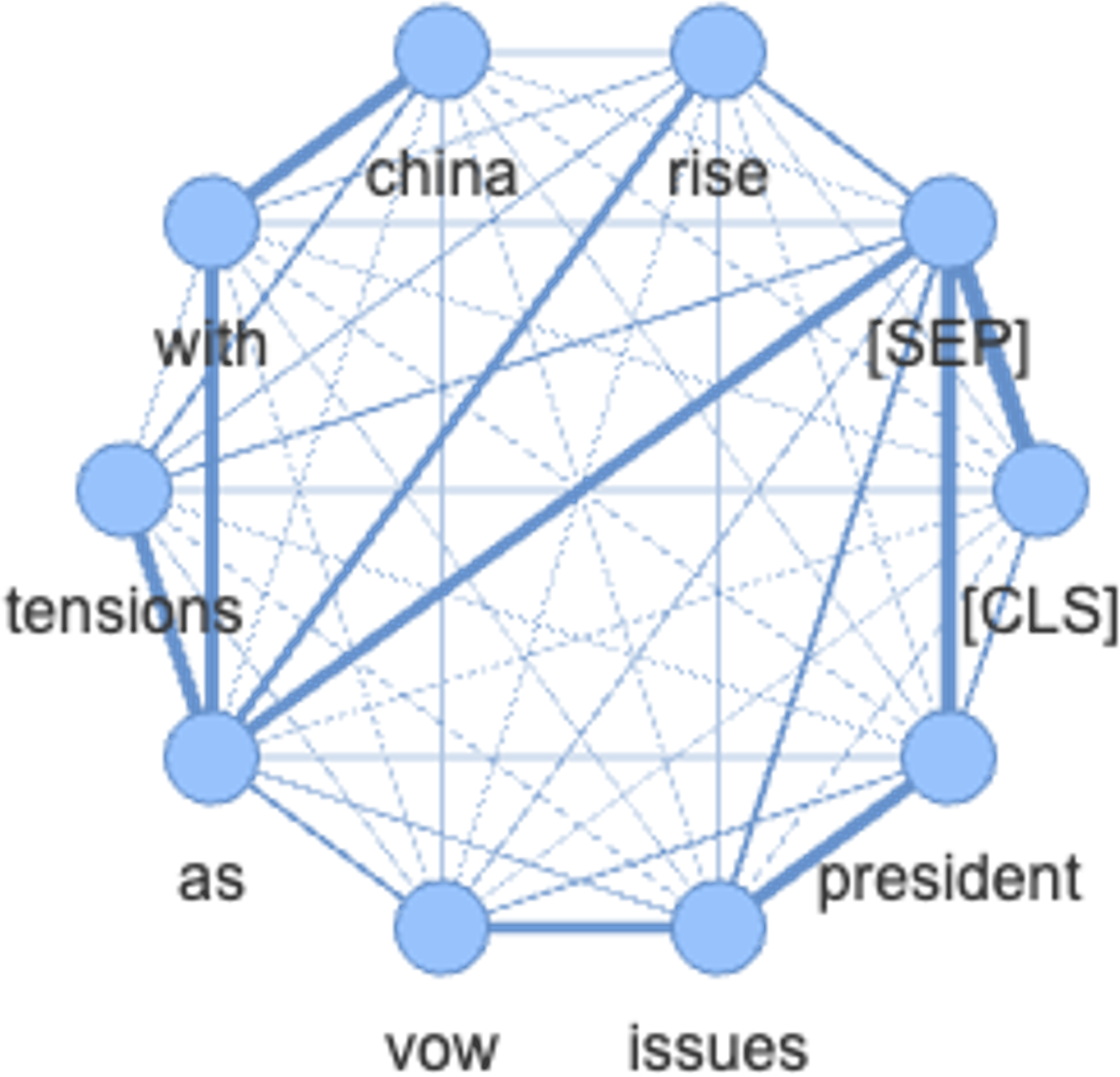}
        \caption{Undirected attention graph (Layer 7; Head 10) where edges are proportional to the maximal attention between the two vertices. The edge width represents shorter distances (attention strength)}
        \label{fig1c}
    \end{subfigure}
   \caption{Process of transforming an attention map to an attention graph (one per attention head)}
   \label{fig1}
\end{figure}
Attention maps play a crucial role in our methodology as they form the basis for extracting topological features used in our OOD detection. An attention map $W^{attn}$ is a $n \times n$-dimensional matrix where each entry represents the attention weight between two tokens. Each element $w_{ij}^{attn}$ can be interpreted as the level of ‘attention’ token $i$ pays to token $j$ in the input sequence during the encoding process. The higher the weight the stronger the relation between two tokens. They are non-negative and the attention weights of a token sum up to one (i.e. $\sum_{j=1}^n w_{ij}^{attn} = 1$ for all $i=1,...,n$.).

To generate topological features from an attention map, we first convert it into an attention graph following the approach of \citet{perez2022topological}. Given an attention matrix $W^{attn}$, we create an undirected weighted graph where the vertices represent the tokens of the input text $x$, and the weights are determined by the attention weights in the corresponding attention map. To emphasise the important relationships and reduce noise, we calculate the distance between vertices as $1-\max(w^{ij}, w^{ji})$. The distance calculation reflects the inverse of the maximum attention weight between two tokens, ensuring the relationship is symmetric and the strong relationships result in smaller distances. To prevent the formation of self-loops, all diagonals in the adjacency matrix are set to 0. \autoref{fig1} shows an example of constructing the attention graph for an attention map. 

\subsection{Persistent Homology}
\label{sec:persistent-homology}

\begin{figure}[ht!]
    \centering
    \includegraphics[width=0.4\textwidth]{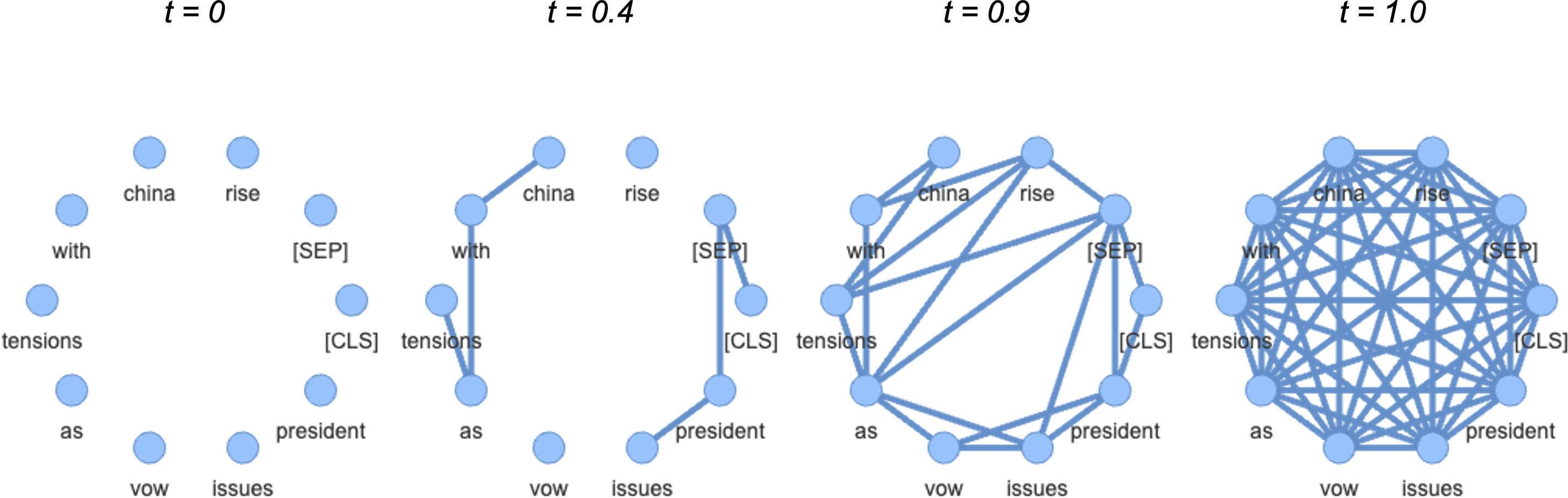}
    \caption{Filtration process for the attention graph (Layer 7; Head 10) where edges with shorter distances below a threshold are added first, gradually connection the nodes until a complete graph is formed}
    \label{fig2}
\end{figure}

The constructed attention graphs from the attention heads contain the structure and relationships we need to extract topological features. To encode the topological information provided by the attention graph, we use a filtration process to generate a persistence diagram. Filtration in TDA is a systematic process where a topological space is progressively constructed across varying scales to analyse the emergence, persistence and disappearance of simplicial complexes, such as connected components, holes, or voids. 

We apply one of the most widely used types of filtration process to the attention graphs, the Vietoris-Rips filtration. This process starts with only the vertices of the graph, considering them as zero-dimensional simplices. Then it adds edges one by one, depending on their weights (i.e. distances). Edges with shorter distances below a threshold are added first, gradually connecting the vertices by increasing the threshold until a complete graph is formed. As edges are added, the filtration process captures the graph’s properties and the relationships between its vertices~\cite{article}. This process is visualised in \autoref{fig2}.

\begin{figure}[tbp!]
    \centering
    \begin{subfigure}[t]{0.2\textwidth}
        \centering
        \includegraphics[width=\linewidth]{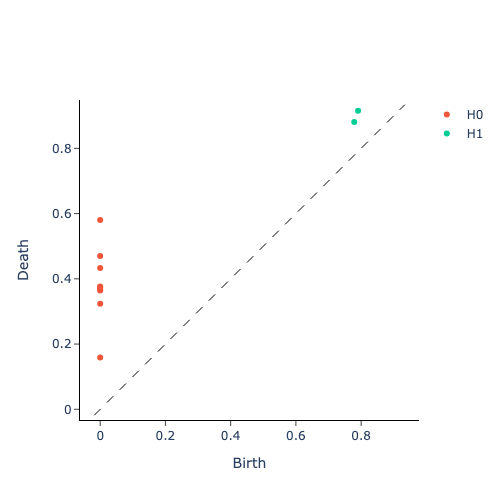}
        \caption{Persistence diagram generated from the filtration process for attention map in Layer 7, Head 10. The set of $H_0$ (red points) represents the birth and death of  `connected components' and the set of $H_1$ (teal points) represents the birth and death of `holes'.}
        \label{fig3a}
    \end{subfigure}
    \quad
    \begin{subfigure}[t]{0.2\textwidth}
        \centering
        \includegraphics[width=\linewidth ]{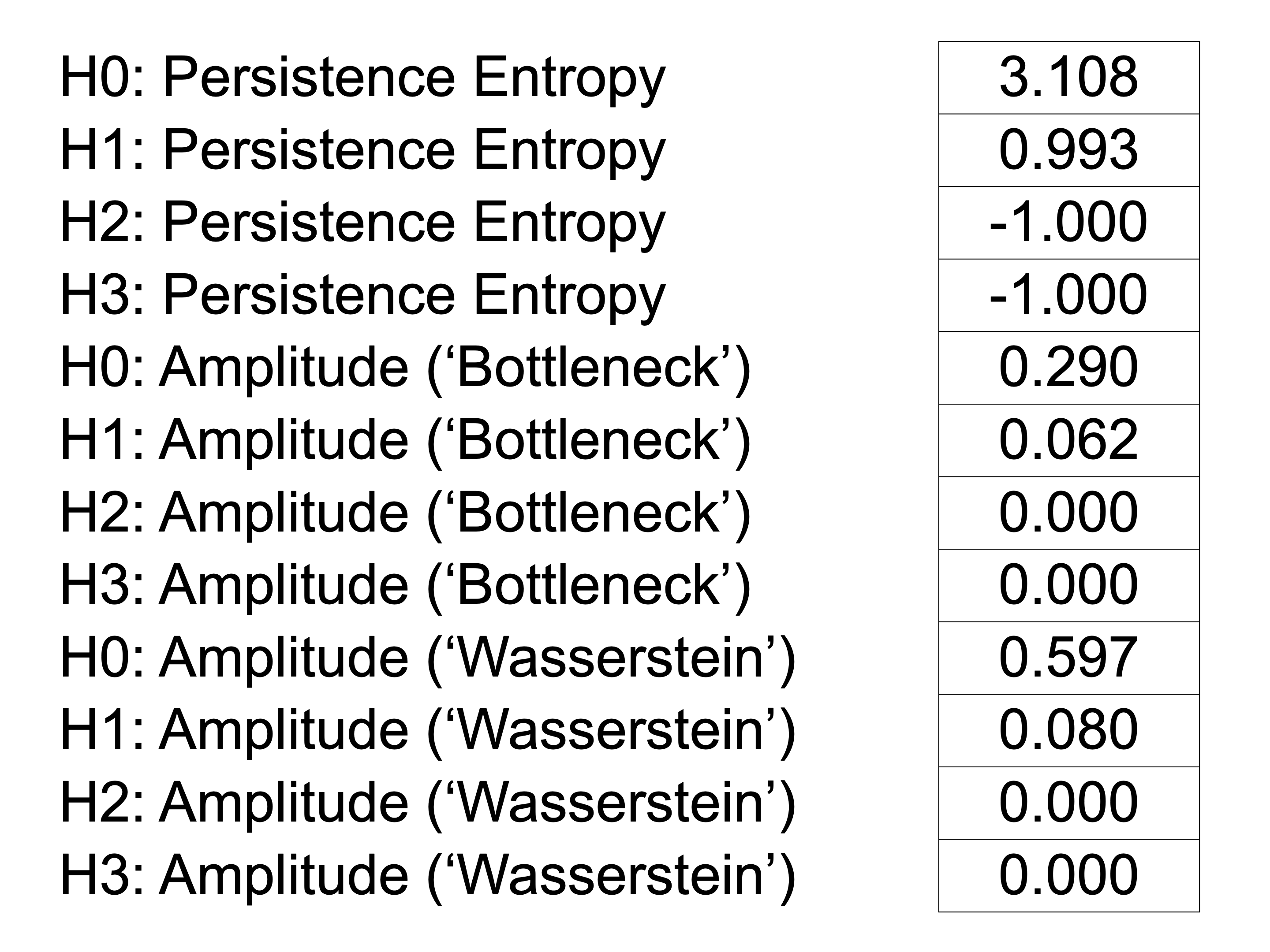}
        \caption{Topological features extracted from the persistence diagram, calculating persistence entropy, and amplitude with `Bottleneck' and `Wassertstein' distances for homology dimensions 0, 1, 2 and 3. (In the case of NaN values, e.g. due to no higher dimensional simplices, we set the persistence entropy feature to -1, as per the default behaviour of Giotto-tda)}
        \label{fig3b}
    \end{subfigure}
    \label{fig3}
    \caption{Example persistence diagram and extracted topological features}
    \end{figure}

To construct a persistence diagram, we keep track of the lifetime of persistence features as the threshold is increased. One can think of 0-dimensional persistent features as connected components, 1-dimensional features as holes and 2-dimensional features as voids (2-dimensional holes) and so on. The birth and death time of a persistence feature is the threshold value at which the feature appeared and disappeared. For example, when the threshold is 0 all 0-dimensional features are born (vertices), and when two vertices $i$ and $j$ are connected at threshold $w^{ij}$, one 0-dimensional feature will disappear. Similarly, a 1-dimensional feature (hole) will appear at the threshold where 3 vertices connect to each other, and disappear when a fourth vertex forms a 2-dimensional simplex (void). The birth and death of all $k$-dimensional simplices are recorded in a persistence diagram. An example persistence diagram is shown in \autoref{fig3a}.

From the persistence diagrams, we extract various topological features to represent the underlying graph’s structure. In our experiments, we focus on the following topological features:
\begin{enumerate}
    \item \textbf{Persistence Entropy}: This feature quantifies the complexity of the persistence diagram as calculated by the Shannon entropy of the persistence values (birth and death), with higher entropy indicating a more complex topology. 
    \item \textbf{Amplitude}: We compute amplitude using two different distance measures: ‘bottleneck’ and ‘Wasserstein’. The amplitude measures the maximum persistence value within the diagram, providing insights into the significance of the topological features. 
\end{enumerate}
We focus on different homology dimensions to capture topological features of varying complexities. In our experiments, we consider homology dimensions [0, 1, 2, 3] to account for different aspects of the attention graph's topology.
We use the Giotto-tda library to generate the persistence diagrams and extract the topological features, as per \autoref{fig3b}.
Both persistence entropy and amplitude features are used in the experiment through concatenating all features into a single feature vector.

\subsection{OOD Scoring Function}
\label{sec:ood-scoring}
Similar to \citet{perez2022topological}, given $h(x)$, a $d$-dimensional representation of an input text $x$, we employ two distance-based methods as the OOD scoring functions:
\begin{enumerate}
    \item \textbf{Mahalanobis distance to the ID class centroids}: the Mahalanobis distance is used to measure the distance between the feature vector $h(x)$ and the class centroids. This distance is based on the covariance matrix of the class features, which is based on the assumption that the data in that class follows a multivariate Gaussian distribution. The OOD score is calculated as follows:
    
\begin{align*}
& S_{\text{Maha}}(x;h; \Sigma; \mu) = \\
& min_{c \in y} (z_x-\mu_c)^T \Sigma^{-1}(z_x-\mu_c)
\end{align*}

where $z_x$ is the standardised feature vector for the input $h(x)$, $\Sigma$ is the covariance matrix of the standardised ID feature vectors and $\mu$ is the set of class mean standardised embeddings. Both $\Sigma$ and $\mu_c$ are extracted from the ID validation set embeddings to account for the inherent distribution of the ID data. The covariance matrix $\Sigma$ captures how the features vary with respect to one another, and $\mu_c$ represents the centroid or average representation of data belonging to class $c$.

    \item \textbf{Euclidean distance to k-nearest ID neighbour}:  We measure the distance between $h(x)$ and the k-nearest ID neighbour’s feature vector from the validation set. Given $h(x)$ and a set of $m$ ID feature vectors $\{h(x_1), h(x_2),...,h(x_m)\}$, the Euclidean distance to the k-nearest ID neighbour is calculated as follows:
    
\[
S_{\text{KNN}}(x;h) = || z_x - z_{x_k}||_2
\]
where $z_x$ and $z_{x_k}$ are the standardised feature vector for the input $h(x)$ and its k-nearest ID sample $h(x_k)$. In our experiments, we set $k=5$. 

\end{enumerate}

\section{Results}
We conduct our experiments using Topological Data Analysis to generate topological feature vectors $h_1(x)$ from attention maps, which are then compared to standard sentence embeddings $h_2(x)$ generated from the $[CLS]$ token of BERT. \autoref{tab1} shows the OOD detection performance of both approaches for three out-of-distribution datasets, using both pre-trained and fine-tuned BERT models.

\begin{figure}
    \centering
    \small
    \begin{tabular}{c|c|c}
        &\textbf{Pre-trained} & \textbf{Fine-tuned} \\ \hline
        \centering \textbf{TDA} & \includegraphics[width=0.2\textwidth]{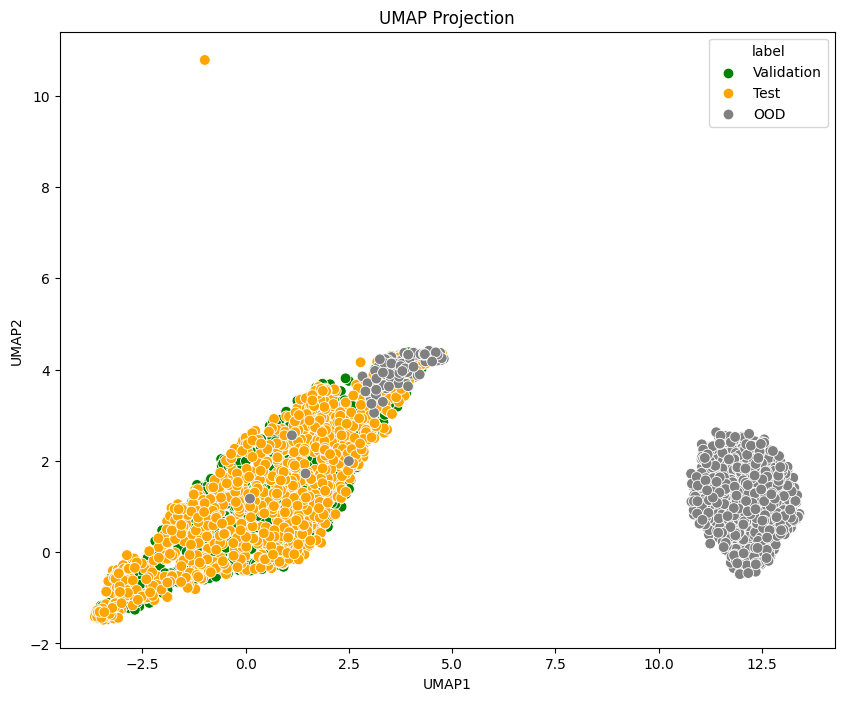} & \includegraphics[width=0.2\textwidth]{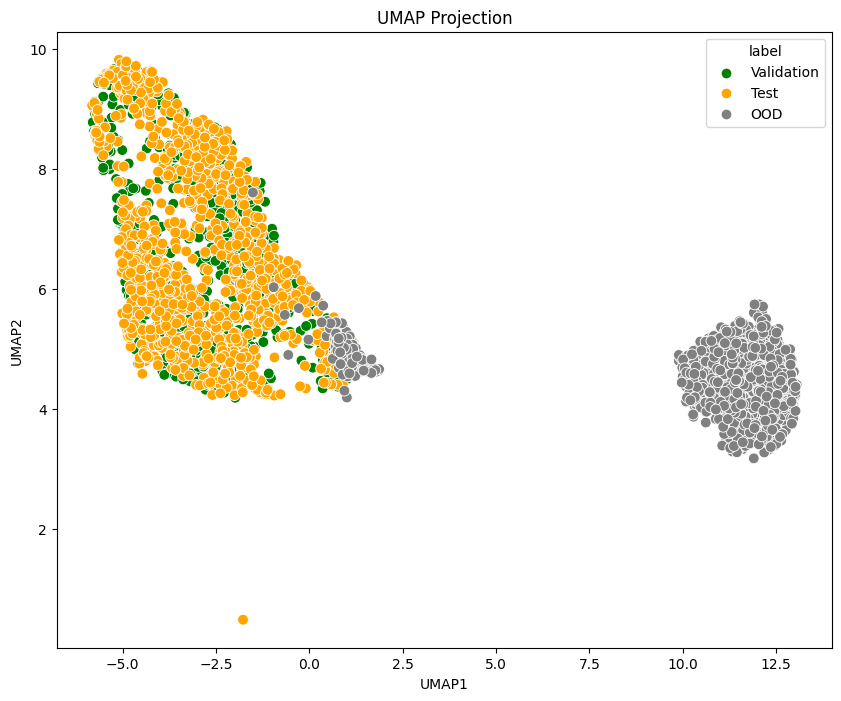} \\ \hline
       \centering \textbf{CLS} & \includegraphics[width=0.2\textwidth]{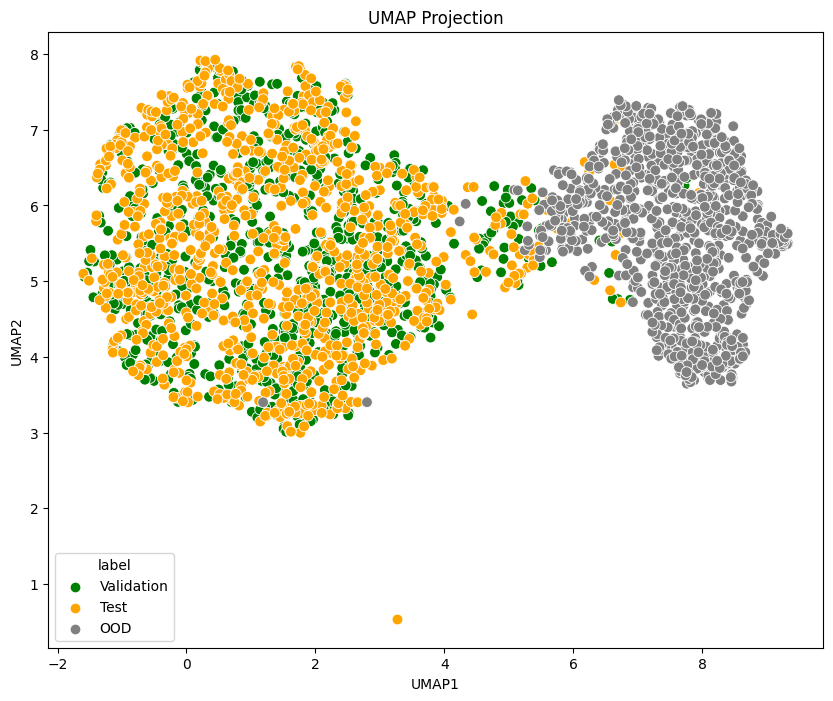} & \includegraphics[width=0.2\textwidth]{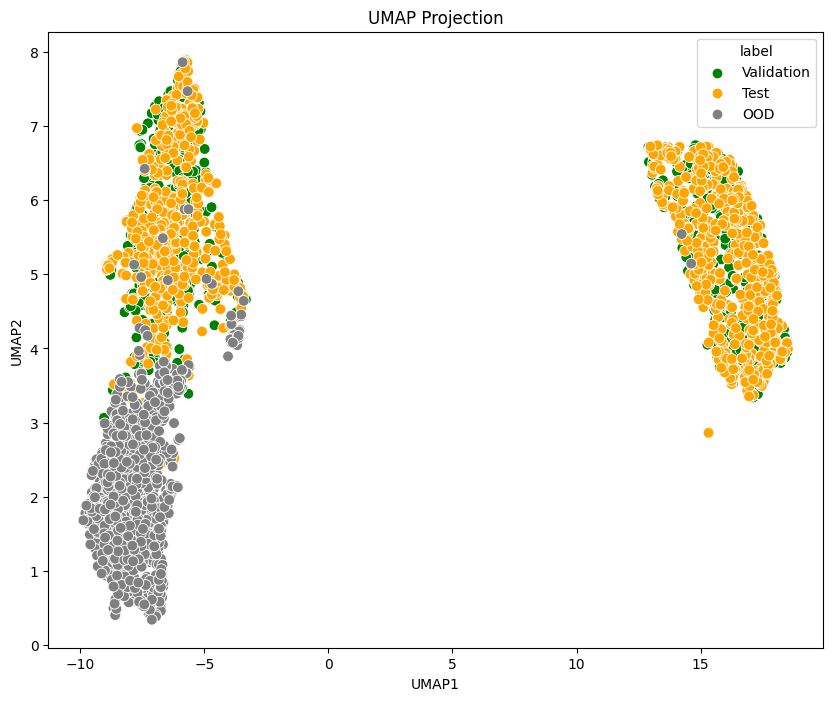} \\
    \end{tabular}
    \caption{The data representations from the TDA and CLS approaches for the far out-of-domain IMDB dataset.}
    \label{fig4}
\end{figure}

\begin{figure}
    \centering
    \small
    \begin{tabular}{c|c|c}
        &\textbf{Pre-trained} & \textbf{Fine-tuned} \\ \hline
        \centering \textbf{TDA} & \includegraphics[width=0.2\textwidth]{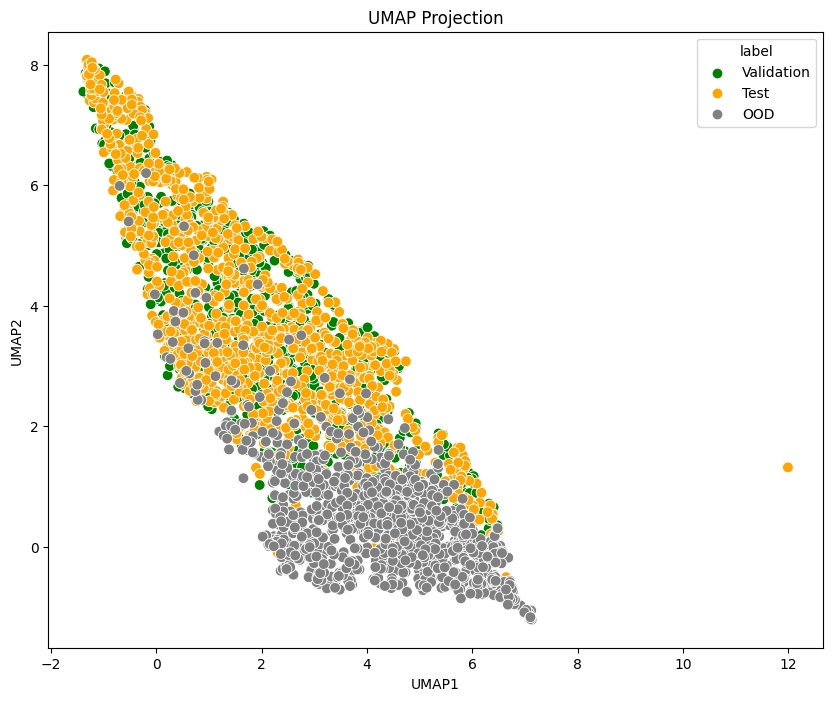} & \includegraphics[width=0.2\textwidth]{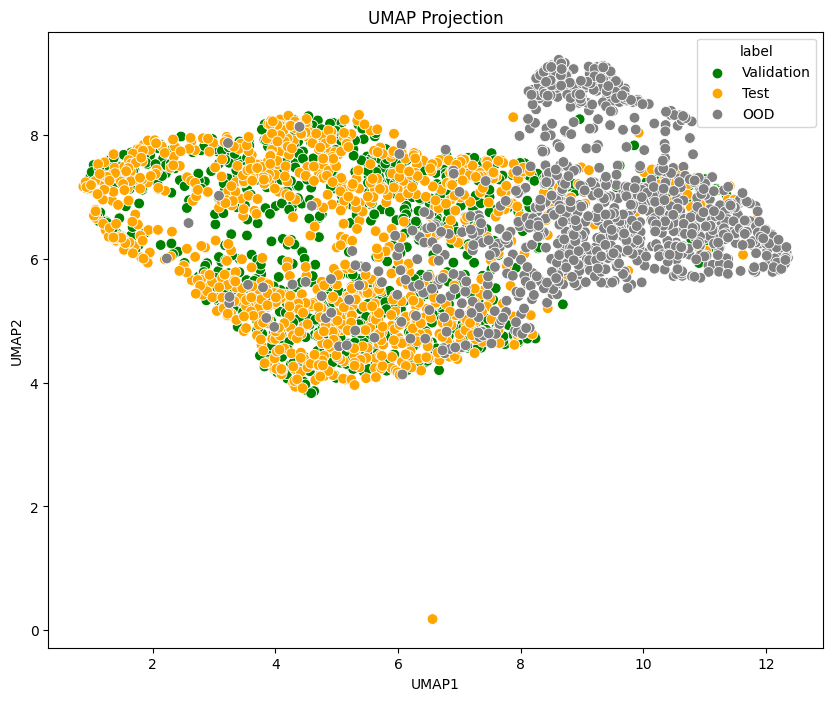} \\ \hline
       \centering \textbf{CLS} & \includegraphics[width=0.2\textwidth]{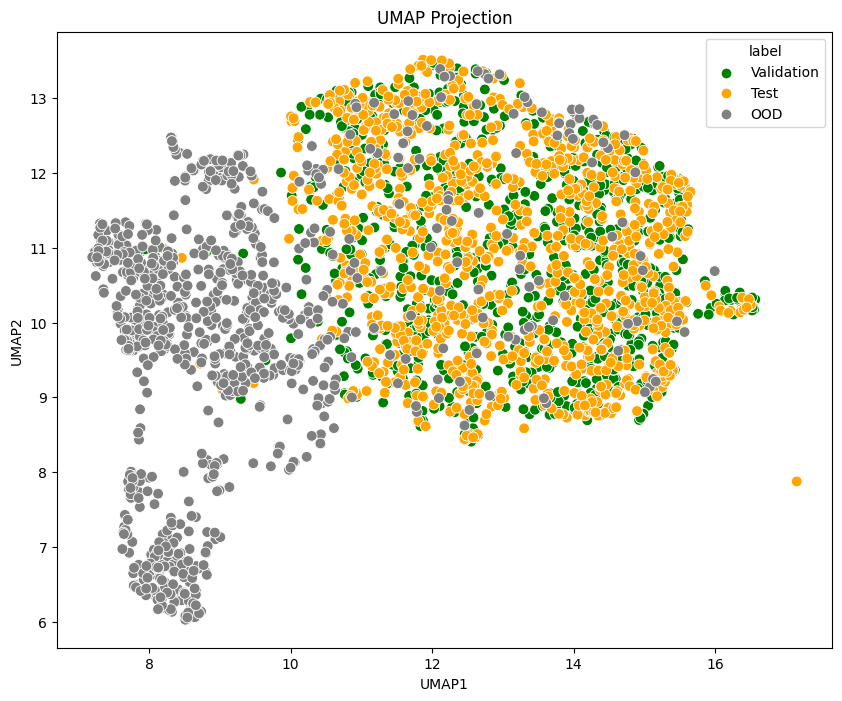} & \includegraphics[width=0.2\textwidth]{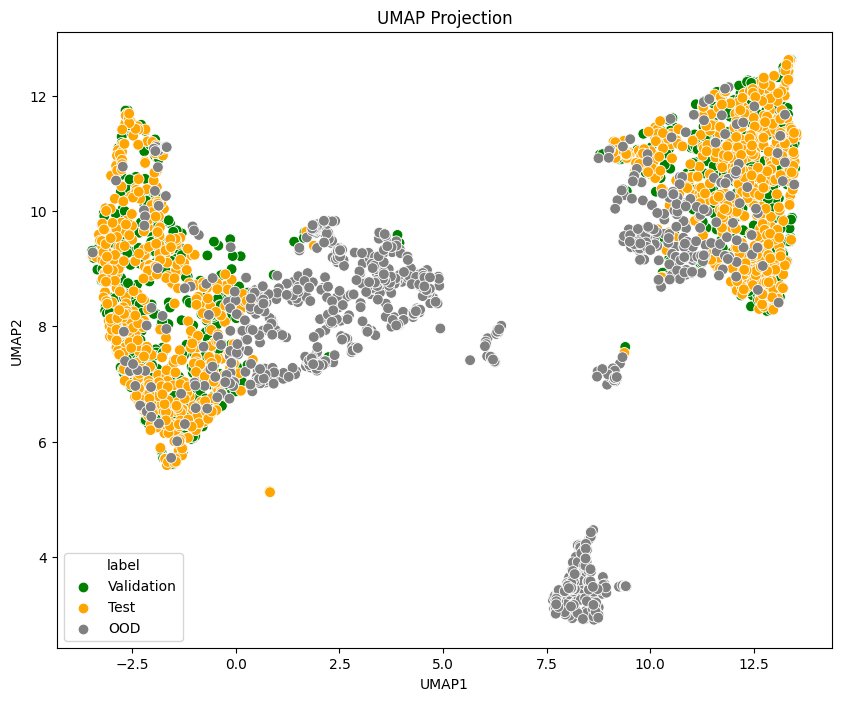} \\
    \end{tabular}
    \caption{The data representations from the TDA and CLS approaches for the near out-of-domain CNN/Dailymail dataset.}
    \label{fig5}
\end{figure}

\begin{figure}
    \centering
    \small
    \setlength{\tabcolsep}{3pt} 
    \begin{tabular}{c|c|c}
        & \textbf{Pre-trained} & \textbf{Fine-tuned} \\ \hline
        \textbf{TDA} & \includegraphics[width=0.45\columnwidth]{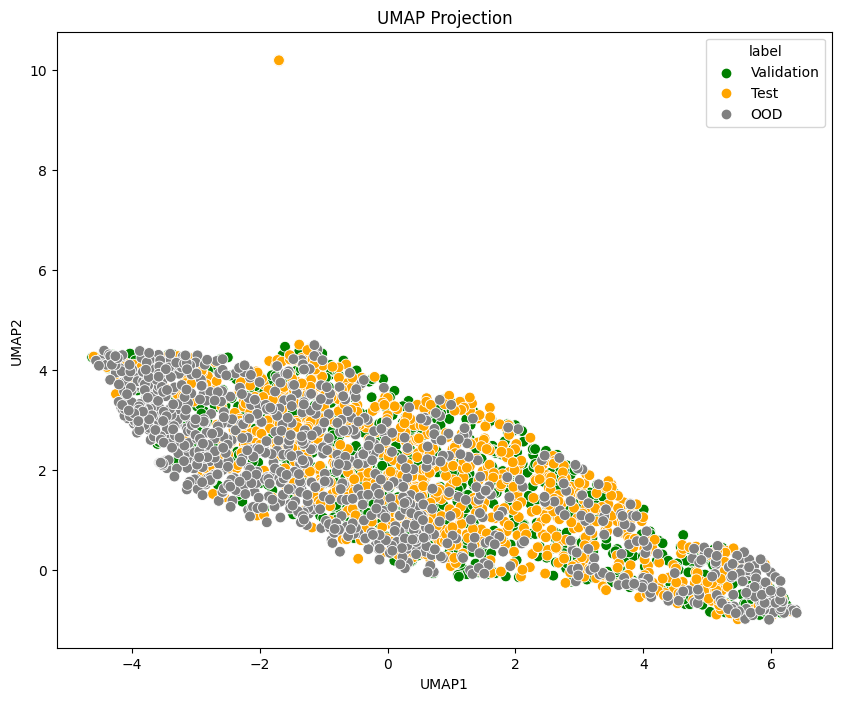} & \includegraphics[width=0.45\columnwidth]{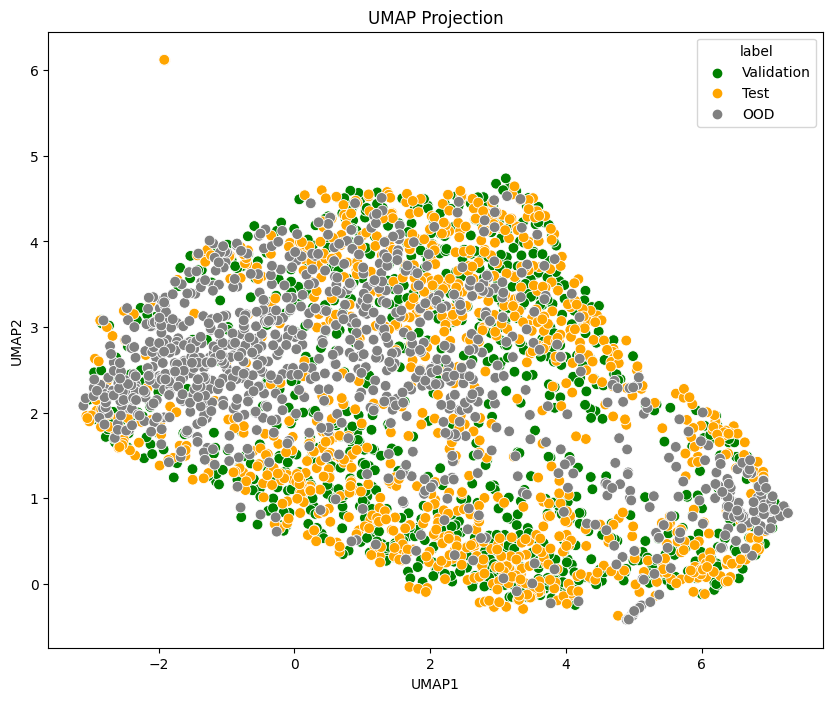} \\ \hline
        \textbf{CLS} & \includegraphics[width=0.45\columnwidth]{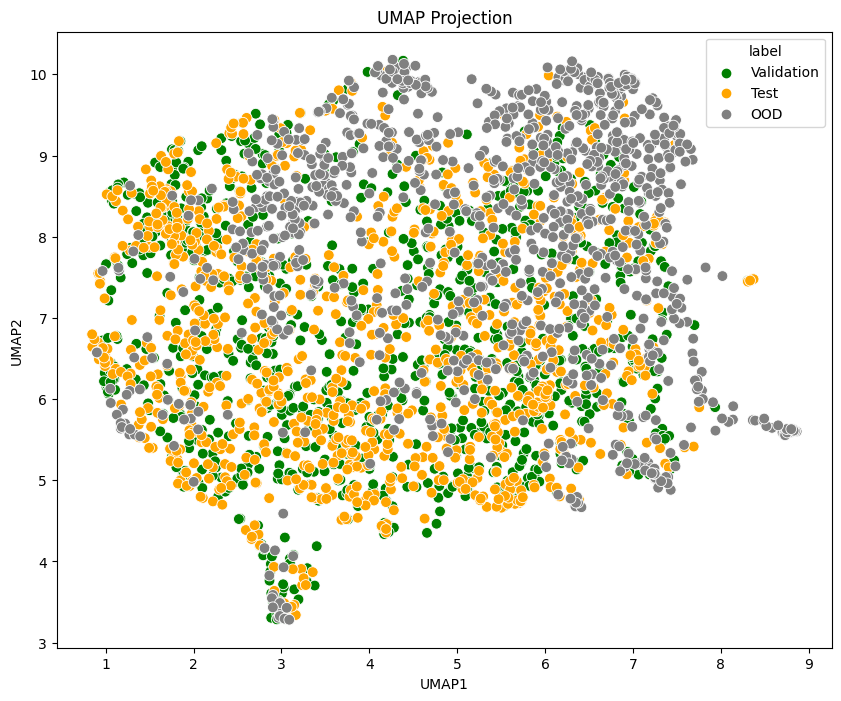} & \includegraphics[width=0.45\columnwidth]{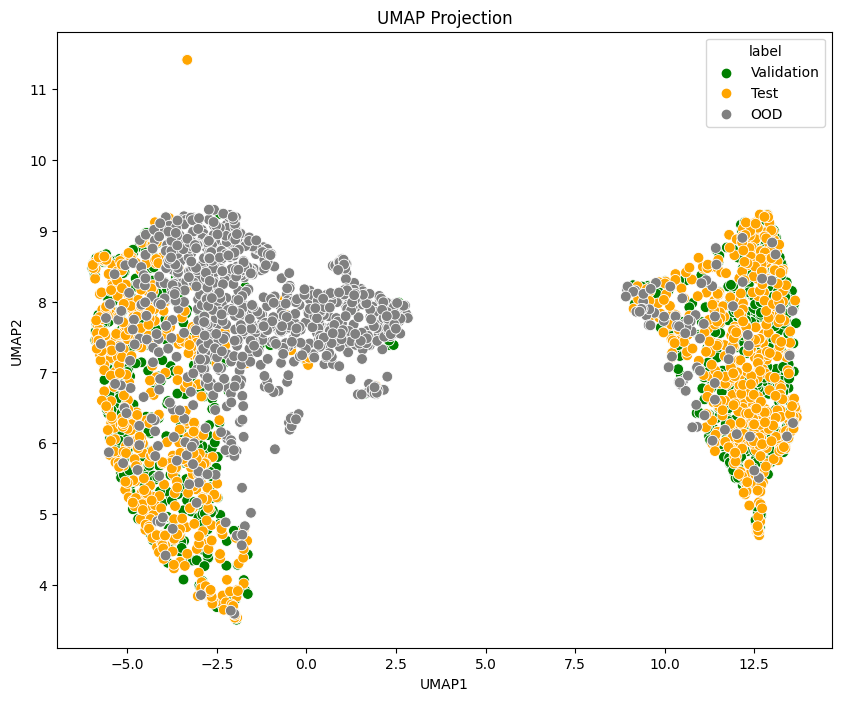} \\
    \end{tabular}
    \caption{The data representations from the TDA and CLS approaches for the same-domain News-Category (Business) dataset.}
    \label{fig6}
\end{figure}

\begin{table*}[htbp]
    \centering
    \small 
    \begin{tabularx}{\textwidth}{l|l|*{8}{>{\centering\arraybackslash}X}}
    \toprule
     & & \multicolumn{4}{c|}{Pre-trained model} & \multicolumn{4}{c}{Fine-tuned model} \\
     & & \multicolumn{2}{c}{KNN} & \multicolumn{2}{c|}{MAHA} & \multicolumn{2}{c}{KNN} & \multicolumn{2}{c}{MAHA} \\
     & & AUROC $\uparrow$ & FPR95 $\downarrow$ & AUROC $\uparrow$ & FPR95 $\downarrow$ & AUROC $\uparrow$ & FPR95 $\downarrow$ & AUROC $\uparrow$ & FPR95 $\downarrow$ \\
     \midrule
    \multirow{2}{*}{IMDB} & TDA & \textbf{0.940} & \textbf{0.090} & \textbf{0.940} & 0.112 & \textbf{0.958} & \textbf{0.084} & 0.950 & 0.124 \\
     & CLS & 0.680 & 0.875 & 0.799 & 0.704 & 0.771 & 0.916 & 0.814 & 0.852 \\
     \midrule
    \multirow{2}{*}{CNN/Dailymail} & TDA & 0.572 & 0.890 & 0.563 & 0.908 & 0.551 & 0.909 & 0.521 & 0.927 \\
     & CLS & 0.875 & 0.591 & \textbf{0.897} & \textbf{0.445} & 0.947 & 0.215 & \textbf{0.949} & \textbf{0.208} \\
     \midrule
    \multirow{2}{*}{News-Category (Business)} & TDA & 0.527 & 0.929 & 0.543 & 0.921 & 0.570 & 0.923 & 0.568 & 0.925 \\
     & CLS & 0.580 & 0.921 & \textbf{0.638} & \textbf{0.878} & 0.884 & 0.431 & \textbf{0.885} & \textbf{0.424} \\
    \bottomrule
    \end{tabularx}
    \caption{Comparison of the performance of our scoring functions on all three out-of-distribution datasets using both pre-trained and fine-tuned models.}
    \label{tab1}
\end{table*}
For visualisation purposes, we use UMAP projections of the in-distribution (validation and test sets) and out-of-distribution data points in the corresponding feature space. \autoref{fig4}, \autoref{fig5}, and \autoref{fig6} show the data representations from the TDA and CLS approaches for the far out-of-domain dataset (IMDB), near out-of-domain dataset (CNN/Dailymail) and the same-domain dataset (business news-category), respectively.

The results demonstrate that the TDA-based approach consistently outperforms the CLS embeddings in detecting OOD samples in the IMDB dataset from both the pre-trained and fine-tuned models. OOD detection using TDA can detect IMDB review samples with 8-9\% FPR95, in stark contrast to the 87-91\% FPR95 exhibited by CLS embeddings. As seen in \autoref{fig4}, the TDA feature vectors project the data into well-separated and compact clusters, which explains its superior performance.

The TDA approach was less effective than the CLS approach at detecting OOD samples from the near out-of-domain CNN/Dailymail dataset. Even though the data visualisation in~\autoref{fig5} shows that TDA was able to cluster OOD samples together, the cluster was not distant enough from ID samples, rendering both distance-based OOD detection methods less effective. 

For same-domain datasets (news-category), both approaches struggled to detect OOD samples. As seen in~\autoref{fig6}, when both ID and OOD data are from the same domain, their feature vectors are highly overlapping, although fine-tuning seems to provide stronger separability between ID and OOD data for the CLS approach.

\section{Discussion}
From our experiments, we showed that the TDA approach outperforms the CLS approach at detecting far out-of-domain OOD samples like those in the IMDB dataset. Yet, its effectiveness deteriorates with near out-of-domain (CNN/Dailymail) or same-domain (business news-category) datasets. To understand why, we looked at the samples that each approach thrived and struggled with, and we highlight three observations:

\textbf{(1) The TDA approach accentuates features associated with textual flow or grammatical structures rather than lexical semantics}, consistent the findings of \citet{DBLP:journals/bdcc/DengD22} and \citet{DBLP:journals/corr/abs-2109-04825}. For example, TDA was adept at identifying OOD samples that are structurally unique in the IMDB dataset, as the most confident OOD samples detected were:
\begin{itemize}
    \item \textit{‘OK...i have seen just about everything....and some are considered classics that shouldn't be ( like all those Halloween movies that suck crap or even Steven king junk).......and some are considered just OK that are really great.....( like carnival of souls )........and then some are just plain ignored............like ( evil ed ) […]’}
    \item \textit{‘Time line of the film: * Laugh * Laugh * Laugh * Smirk * Smirk * Yawn * Look at watch * walk out * remember funny parts at the beginning * smirk < br / > <br /> […]’}
\end{itemize}

In contrast, TDA struggled with detecting CNN/Dailymail OOD samples as they have similar sentence structures and length to the ID samples, even if they are semantically unrelated. \autoref{tab2} shows the samples with the least confident OOD score from the CNN/Dailymail dataset, and their nearest ID neighbour.

\begin{table}[t]
    \centering
    \small 
    \begin{tabularx}{\linewidth}{X|X}
    \toprule
    \textbf{CNN/Dailymail sample} & \textbf{Nearest ID neighbour} \\
    \midrule
    Footage showed an unusual 'apocalyptic' dust storm hitting Belarus. China has suffered four massive sandstorms since the start of the year. Half of dust in atmosphere today is due to human activity, said Nasa. & Trump's Proposed Cuts To Foreign Food Aid Are Proving Unpopular. The president might see zeroed-out funding for foreign food aid as "putting America first," but members of Congress clearly disagree. \\
    \midrule
    Video posted by YouTube user Richard Stewart showing a Porsche Cayman flying out of control. Police cited unidentified driver for the crash. Car reportedly wrecked and needed to be towed from the scene. & Trump Signs Larry Nassar-Inspired Sexual Assault Bill Behind Closed Doors. The president quietly signed the bill the week after two White House staffers resigned amid allegations of domestic violence. \\
    \bottomrule
    \end{tabularx}
    \caption{Least confident OOD samples from the CNN/Dailymail dataset and their nearest ID neighbours, from the TDA approach using the pre-trained BERT model}
    \label{tab2}
\end{table}

\textbf{(2) CLS embeddings are sensitive to the semantic and contextual meaning of the samples, regardless of sentence structure}. This explains why this approach struggled with OOD detection from IMDB reviews, as it often classified IMDB movie reviews as in-distribution due to their semantic similarities with the entertainment news articles from the ID dataset, especially those related to movies. A closer look at the IMDB samples with smallest OOD score from the CLS embeddings in~\autoref{tab3} exemplifies this insight, identifying ID samples of similar topic as nearest neighbours even though they are clearly from different domains.

\begin{table}[h]
    \centering
    \small 
    \begin{tabularx}{\linewidth}{X|X}
    \toprule
    \textbf{IMDB review sample} & \textbf{Nearest ID neighbour} \\
    \midrule
    '[...] I would spend good, hard-earned cash money to see it again on DVD. And as long as we're requesting Smart Series That Never Got a Chance...How about DVD releases of Maximum Bob (another well written, odd duck show with a delightful cast of characters.) [...]' & DVDs: Great Blimp, Badlands, Buster Keaton \& More. Let's catch up with some reissues of classic -- and not so classic -- movies, with a few documentaries tossed in at the end for good measure. \\
    \midrule
    '[...] I am generally not a fan of Zeta-Jones but even I must admit that Kate is STUNNING in this movie. [...]' & How ‘Erin Brockovich’ Became One Of The Most Rewatchable Movies Ever Made. Julia Roberts gives the best performance of her career, aided by a sassy Susannah Grant script full of one-liners. \\
    \bottomrule
    \end{tabularx}
    \caption{Least confident OOD samples from the IMDB dataset and their nearest ID neighbours, from the CLS approach using the pre-trained BERT model}
    \label{tab3}
\end{table}

\textbf{(3) Fine-tuning has improved performance of CLS embeddings for near or same-domain shifts, but shows no significant benefit for TDA}. Fine-tuning induces a model to divide a single domain cluster into class clusters, as highlighted by \citet{uppaal-etal-2023-fine}. For CNN/Dailymail and Business news OOD datasets, this is beneficial for the CLS approach as it learns to better distinguish topics. However, fine-tuning made the CLS embeddings of IMDB movie reviews appear even more similar to entertainment news, deteriorating OOD performance. 

For the TDA approach, fine-tuning did not present any considerable benefits. This can be partly attributed to observation (1) that TDA primarily captures structural differences, and fine-tuning, which is driven by semantics, does not significantly alter the topological representation.

\section{Conclusion}
In this paper, we explore the capabilities of Topological Data Analysis for identifying Out-of-Distribution samples by leveraging the attention maps derived from BERT, a transformer-based Large Language Model. Our results demonstrate the potential of TDA as an effective tool to capture the structural information of textual data. 

Nevertheless, our experiments also highlighted the intrinsic limitations of TDA-based methods. Predominantly, our TDA method captured the inter-word relations derived from the attention maps, but failed to account for the actual lexical meaning of the text. This distinction suggests that while TDA offers valuable insights into textual structure, a lexical and more holistic understanding of textual data is needed for OOD detection, especially with near or same-domain shifts. 

For future work, it might be worth combining the topological features that capture the structural information of textual data, with those that encode the semantics of text in an ensemble model that might boost our ability to detect OOD samples. In addition, there is an opportunity to investigate the effectiveness of TDA in other NLP tasks where the textual structure might be important. 

\begin{acknowledgments}
The research was supported by a National Intelligence Postdoctoral Grant (NIPG-2021-006).
\end{acknowledgments}


\bibliography{main}

\end{document}